\documentclass[pmlr]{jmlr}%

\usepackage{longtable}%

\usepackage{booktabs}

\usepackage[load-configurations=version-1]{siunitx} %

\usepackage{amsmath}
\DeclareMathOperator*{\argmax}{arg\,max}

\usepackage{tikz,pgfplots,pgfplotstable,pgf-pie}
\usepackage{placeins}
\usepackage{xcolor}

\theorembodyfont{\upshape}
\theoremheaderfont{\scshape}
\theorempostheader{:}
\theoremsep{\newline}

\usepackage[utf8]{inputenc} %
\usepackage[T1]{fontenc}    %
\usepackage{booktabs}       %
\usepackage{nicefrac}       %
\usepackage{microtype}      %
\usepackage{graphics,graphicx,color}
\usepackage{wrapfig}

\usepackage{algorithm}
\usepackage{algorithmic}
\usepackage{enumitem}

\usepackage{amsfonts}       %
\usepackage{amsmath}       %
\usepackage{amssymb}
\usepackage[export]{adjustbox}
\usepackage{multirow} %

\usepackage{siunitx} %
\usepackage{makecell} %

\newcommand{\tabref}[1]{Table~\ref{#1}}

\usepackage{xspace}
\makeatletter
\DeclareRobustCommand\onedot{\futurelet\@let@token\@onedot}
\def\@onedot{\ifx\@let@token.\else.\null\fi\xspace}

\makeatother

\usepackage{xr-hyper}
\makeatletter
\newcommand*{\addFileDependency}[1]{%
  \typeout{(#1)}
  \@addtofilelist{#1}
  \IfFileExists{#1}{}{\typeout{No file #1.}}
}
\makeatother

\definecolor{ourblue}{rgb}{0.368,0.507,0.71}
\definecolor{ourorange}{rgb}{0.881,0.611,0.142}
\definecolor{ourgreen}{rgb}{0.56,0.692,0.195}
\definecolor{ourred}{rgb}{0.923,0.386,0.209}
\definecolor{ourviolet}{rgb}{0.528,0.471,0.701}
\definecolor{ourbrown}{rgb}{0.772,0.432,0.102}
\definecolor{ourlightblue}{rgb}{0.364,0.619,0.782}
\definecolor{ourdarkgreen}{rgb}{0.572,0.586,0.}

\definecolor{ourcyan2}{rgb}{0.125,0.722,0.804}
\definecolor{ourred2}{rgb}{0.863,0.184,0.047}
\definecolor{ouryellow2}{cmyk}{0,0.16,1.0,0.07}
\definecolor{ourviolet2}{cmyk}{0.55,0.56,0,0.47}
\definecolor{ourorange2}{cmyk}{0,0.46,0.89,0.11}

\def\figref#1{Fig.~\ref{#1}}

\newcommand{\DPushSimExp}{Sim-Expert\xspace}

\newcommand{\DPushRealExp}{Real-Expert\xspace}

\newcommand{\DLiftSimExp}{Sim-Expert\xspace}

\newcommand{\DLiftRealExp}{Real-Expert\xspace}

\renewcommand{\paragraph}[1]{\par{\bf #1}\ \ }

\jmlrvolume{}
\jmlryear{}
\jmlrworkshop{}

\title[Real Robot Challenge 2022]{Real Robot Challenge 2022:\\ Learning Dexterous Manipulation\\ from Offline Data in the Real World}

  \author{
  \Name{Nico Gürtler}$^1$, 
  \Name{Felix Widmaier}$^1$, 
  \Name{Cansu Sancaktar}$^1$, 
  \Name{Sebastian Blaes}$^1$, 
  \Name{Pavel Kolev}$^1$, 
  \Name{Stefan Bauer}$^2$, 
  \Name{Manuel W\"uthrich}$^3$, 
  \Name{Markus Wulfmeier}$^4$, 
  \Name{Martin Riedmiller}$^4$, 
  \Name{Arthur Allshire}$^5$, 
  \Name{Qiang Wang}$^{\dagger6}$, \Name{Robert McCarthy}$^{\dagger6}$, \Name{Hangyeol Kim}$^{\dagger7}$, \Name{Jongchan Baek}$^{\dagger7}$, \Name{Wookyong Kwon}$^{\dagger8}$, \Name{Shanliang Qian}$^{\dagger}$, \Name{Yasunori Toshimitsu}$^{\dagger9}$, \Name{Mike Yan Michelis}$^{\dagger9}$, \Name{Amirhossein Kazemipour}$^{\dagger9}$,
\Name{Arman Raayatsanati}$^{\dagger9}$, \Name{Hehui Zheng}$^{\dagger9}$, \Name{Barnabas Gavin Cangan}$^{\dagger9}$,
  \Name{Bernhard Sch\"olkopf}$^1$, \and 
  \Name{Georg Martius}$^1$ \\
  \addr $^1$Max Planck Institute for Intelligent Systems, Tübingen, Germany\\
  \addr $^2$Helmholtz and TU Munich\\
  \addr $^3$Harvard University\\
  \addr $^4$DeepMind\\
  \addr $^5$University of Toronto\\
  \addr $^6$University College Dublin\\
  \addr $^7$Pohang University of Science and Technology\\
  \addr $^8$Electronics and Telecommunications Research Institute\\
  \addr $^9$ETH Zurich\\
  \addr $^\dagger$Competition participants
   }

\editor{Editor's name}

\begin{document}

\maketitle

\begin{abstract}
Experimentation on real robots is demanding in terms of time and costs. 
For this reason, a large part of the reinforcement learning (RL) community uses simulators to develop and benchmark algorithms. 
However, insights gained in simulation do not necessarily translate to real robots, in particular for tasks involving complex interactions with the environment.
The \emph{Real Robot Challenge 2022}\footnote{\url{https://real-robot-challenge.com/}} therefore served as a bridge between the RL and robotics communities by allowing participants 
to experiment remotely with a \emph{real} robot -- as easily as in simulation.

In the last years, offline reinforcement learning has matured into a promising paradigm for learning from pre-collected datasets, alleviating the reliance on expensive online interactions. 
We therefore asked the participants to learn two dexterous manipulation tasks involving pushing, grasping, and in-hand orientation from provided real-robot datasets.
An extensive software documentation and an initial stage based on a simulation of the real set-up made the competition particularly accessible.
By giving each team plenty of access budget to evaluate their offline-learned policies on a cluster of seven identical real TriFinger platforms, we organized an exciting competition for machine learners and roboticists alike.  

In this work we state the rules of the competition, present the methods used by the winning teams and compare their results with a benchmark of state-of-the-art offline RL algorithms on the challenge datasets.

\end{abstract}

\newpage
\section{Introduction} %

Robots have the potential to help humans in many tasks provided that they are adaptive and versatile.  
Learning methods are a promising route to creating such flexible control strategies, as they can learn to cope with the complexities of the real world. 
Indeed, reinforcement learning (RL) approaches have recently achieved good performance in challenging robotics tasks \citep{Kalashnikovetal2018:QTOpt, OpenAI2019-gt, rudin22a}.
However, training such policies requires either a large number of expensive and potentially unsafe environment interactions \citep{dulac-arnold2020} or a good simulator.
The field of offline RL \citep{lange2012batch, LKTF20, PMC22} therefore aims to learn from pre-existing datasets without the need for online interactions.

This paradigm could potentially have a transformative effect on robotics similar to the impact of large datasets in supervised learning. Yet, as the experiments on real robots are costly and time consuming, the offline RL community mostly benchmarks their algorithms in simulated environments \citep{FuKNTL20}.
It is, however, not clear to which extent results obtained in simulation transfer to the real world with its noisy and delayed observations and complex dynamics.

To fill this gap, we organized the \emph{Real Robot Challenge 2022} which was hosted at NeurIPS 2022. We asked the community to learn two dexterous manipulation tasks from pre-collected real-robot datasets we provided, using either offline RL or imitation learning. We chose dexterous manipulation as a challenge as it is a fundamental building block for more complex tasks and a challenging research topic in its own right. Participants could evaluate their solutions remotely by submitting them to a cluster of TriFinger robots \citep{trifinger} hosted at the Max Planck Institute for Intelligent Systems, Tübingen, Germany. 

In the rest of this paper, we describe the challenge in detail in section \ref{sec:challenge}, explain the data collection in \ref{sec:data_collection}, present the baselines and top submissions in section \ref{sec:methods} and discuss the results in section \ref{sec:results}. Finally, we summarize takeaways from the competition in section \ref{sec:takeaways}.

\section{Related Work} %
There have been numerous reinforcement learning competitions for continuous control problems at top machine learning conferences.
However, almost all of them exclusively focused on scenarios in simulation. 
For instance, in the NeurIPS competition track from 2019\footnote{\url{https://nips.cc/Conferences/2019/CompetitionTrack}} and 2020\footnote{\url{https://neurips.cc/Conferences/2020/CompetitionTrack}}, the only competition involving real robots was AI Driving Olympics 3 and 5, which provided a toy environment aiming to replicate a real system via miniaturized self-driving cars~\citep{censi2019ai}. However, it did not include the highly non-linear behavior of contacts which are ubiquitous in manipulation and difficult to learn.
All other robotic challenges (e.g.\ \emph{REAL}~\citep{cartoni2020real}, \emph{MineRL}~\citep{kanervisto2022minerl} and \emph{Learn to Move}~\citep{song2021deep}) were restricted to simulations.
Unfortunately, the policies learned in simulation often do not transfer to the real world.

The \emph{Real Robot Challenge II}~\citep{rrc2020_and_2021}, hosted last year at NeurIPS, was the first challenge in the NeurIPS competition track that is geared towards learning methods for control on real robots in a fully remote setup. However, in the previous instantiations of the \emph{Real Robot Challenge}, there were no restrictions on the algorithms used for controlling the robot.

In this year's competition, we exclusively focused on the offline RL paradigm. 
The goal of offline RL is to learn effective policies from large and diverse datasets covering a sufficient amount of expert transitions without additional online interaction~\citep{LKTF20}. Although several algorithmic advances have been proposed in offline RL in recent years, a standardized benchmark of real-word robotic data has not been established yet. As featured in \cite{Mandlekar2021WhatMattersRL}, there exist small real-world datasets with human demonstrations for a robot arm with a gripper (using operational space control). However, a dataset sufficiently large for offline RL with low-level control to solve more challenging manipulation tasks had been missing. In our challenge, we have provided one such benchmarking dataset that can easily be evaluated remotely on a real-robot platform. This benchmark dataset has also been featured in our concurrent work \citep{guertler2023benchmarking}.

\section{Challenge}
\label{sec:challenge}
The goal of the Real Robot Challenge 2022 was to solve manipulation tasks with TriFinger robots by learning solely from pre-recorded datasets, without access to additional online interactions. 

\subsection{Tasks and Stages} %

We considered two dexterous manipulation tasks involving a tracked cube (see \figref{fig:trifingerpro}, right):
\medskip

{\parindent0pt\paragraph{Push} The goal is to push the cube to a target position which is sampled from a uniform distribution on the ground of the arena. The orientation of the cube does not influence the reward in this task.}
\medskip

{\parindent0pt\paragraph{Lift} For the Lift task a target position in the air and a target orientation have to be matched. The target position is sampled up to a height of \SI{10}{cm} such that the desired cube pose does not intersect with the ground. The desired orientation is sampled uniformly.}
\medskip

\noindent The Lift task is significantly more challenging than the Push task as it requires flipping the cube to an approximately correct orientation, acquiring a stable grasp, lifting it to the goal position and turning it in-hand to match the target orientation. If the cube slips from the fingers all progress is usually lost. This renders the Lift task -- together with the noise on the pose estimation of the cube -- quite unforgiving.

We calculate the reward by applying a logistic kernel to the difference between desired and achieved position (for the Push task) or to the differences between the desired and achieved corner points of the cube (for the Lift task) similar to \citet{Allshire21}. This choice results in a smooth falloff of the reward when deviating from the goal and does not require manually balancing the influence of position and orientation. Further details can be found in Appendix \ref{apd:reward-success}.

We divided the challenge into two overlapping stages:
\medskip

{\parindent0pt\paragraph{Pre-stage (July 1 to September 1, 2022):}  } The pre-stage served as an open qualification round in which everybody could participate. The objective was to learn proficient policies for the Push and Lift tasks from provided simulated datasets containing expert trajectories. The submissions were then evaluated in a simulated version of the TriFinger platform. Teams that reached a promising level of performance on both tasks were admitted to the next stage.
\medskip

{\parindent0pt\paragraph{Real-robot stage (August 1 to October 7, 2022):} With the start of the real-robot stage we released four datasets recorded on real TriFinger robots. All qualifying teams were provided with remote access to the robot cluster (see section \ref{sec:submission_system}).} For each task, two policies had to be learned separately from two datasets with different compositions (see section \ref{sec:data_collection}). 

\subsection{Hardware} %

\begin{figure}
    \centering
    \begingroup
    \setlength{\tabcolsep}{2pt}
    \begin{tabular}{crc}
        & & \hspace{-0.12cm}initial state \hspace{0.00cm} final state \\
         \multirow[c]{1}{*}[1.65cm]{\includegraphics[height=0.27\textwidth]{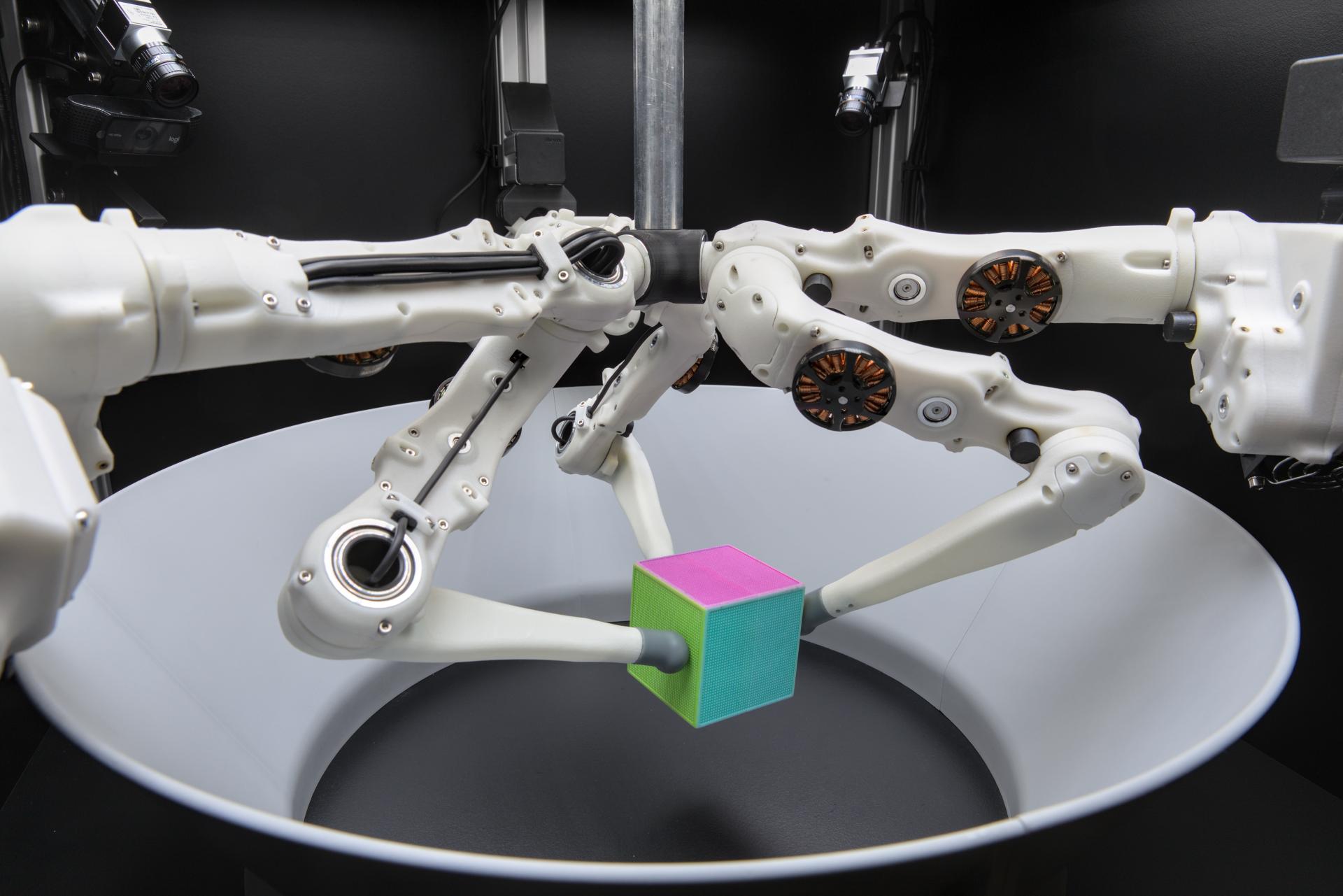}\hspace{0.4cm}} & \rotatebox{90}{\hspace{0.55cm}Push} & \includegraphics[height=0.13\textwidth]{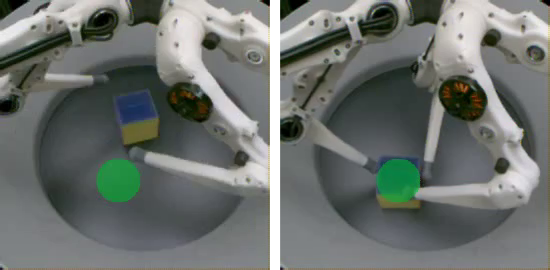}  \\
         & \rotatebox{90}{\hspace{0.75cm}Lift} & \includegraphics[height=0.13\textwidth]{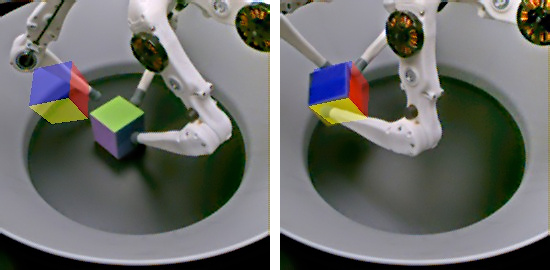}
    \end{tabular}
    \endgroup
    \caption{Left: TriFingerPro robot holding a cube. Right: Examples of initial and final states of successful episodes of the Push and Lift tasks.}
    \label{fig:trifingerpro}
\end{figure}

For the real-robot phase we used the same \emph{TriFingerPro} robot
cluster that was already used in the previous iterations of the
competition \citep{rrc2020_and_2021}.

The \emph{TriFingerPro} robots consist of three fingers that can pick up and manipulate objects within a circular arena (see left side of \figref{fig:trifingerpro}).
They are an enhancement of the open-source
\emph{TriFingerEdu} \citep{trifinger} to make them even more robust and thus
reduce the maintenance effort. The cluster consists of seven such robots
that can be accessed remotely (see section \ref{sec:submission_system}). 
The joints can be torque-controlled through electric
motors. Push sensors on the finger tips can be used to detect contact.
Each robot platform is further equipped with three RGB cameras that are
used to estimate the pose of the manipulated object.

The robots are designed to be able to operate 24/7 without human
supervision. This is made possible by the robust hardware design, several
safety measures in the software (e.g.\ limits on torques or position range
of the joints), automated self-tests after each run and the ability to reset the object position autonomously.

\subsection{Rules} %

The participants were allowed to use any method to learn policies from the provided datasets as long as they complied with the following rules\footnote{The complete list of rules for the competition (including technicalities) is given in Appendix \ref{apd:rules}.}:
\begin{itemize}
    \item Policies had to be learned exclusively from one dataset at a time, i.e., combining datasets or training on additional data (simulated or real) was not allowed.
    \item The training code had to be released under an OSI-approved license and a report describing the method had to be published in a publicly accessible way (e.g. on arXiv).
\end{itemize}
The final ranking of the teams was determined by their submissions in the real-robot stage, as the pre-stage served only as a qualifying round. The main criterion was the score obtained by the submitted policies (see section \ref{sec:evaluating_submissions}). For teams which were closely matched in terms of scores, we also considered the quality of the published report as a secondary criterion for the final ranking.

Due to generous support by DeepMind, we were able to award the following sums as prize money to the highest ranking teams: (1) 2500 USD (2) 1500 USD (3) 1000 USD (split into two times 500 USD as the third rank was shared by two teams with similar results).

\subsection{Submission System} %
\label{sec:submission_system}

In the real-robot stage, participants got access to the submission
system of our robot cluster. This system allows users to remotely submit jobs with their policies to
the cluster, which are then automatically executed on a randomly
selected robot. The resulting data can then be downloaded by the user,
once the job is finished.

Compared to the previous challenges \citep{rrc2020_and_2021}, two major
changes were made:

\begin{enumerate}
\item
  Participants only implemented a policy using a given interface in
  Python. The actual control loop was implemented on the server side and
  could not be modified by the participants (apart from a few
  configuration options like specifying the task).
\item
  Since the task was to learn policies from the given datasets,
  participants were not allowed to collect additional data for the
  training. Therefore the recorded sensor data was not provided to
  participants. They only received the resulting score and related
  statistics as well as a video of their runs.
\end{enumerate}

One job has a runtime of around four minutes, so multiple episodes can
be executed within one job, depending on the task. For the push task
nine episodes are executed, for the lift task (which has a longer
episode length) only six. Between each episode a short ``object reset
trajectory'' is executed to bring the cube back towards the center of
the arena.

\subsection{Evaluating Submissions} %
\label{sec:evaluating_submissions}

\subsubsection*{Pre-stage}

Solutions for the pre-stage were submitted via a form and evaluated in the simulated environment. The same evaluation protocol as in the real-robot stage (see below) was used except for sampling random goals instead of using a fixed sequence.

\subsubsection*{Real-robot stage}

For evaluating the policies of the participants we used the following
procedure:

\begin{enumerate}
\item
  Get the current code of all teams from the submission system.
\item
  Select \(N_R\) robots to be used for the evaluation. For each robot
  generate a list of \(N_G\) random goals for each task/dataset
  combination.
\item
  Run the code of all teams on these goals.
\end{enumerate}

For each episode a score is computed as the cumulative reward over all
time steps. For each task/dataset combination the mean score of the
episodes of all corresponding runs is computed.

The total score for the ranking is then the mean over all task/dataset
combinations. Note that the episode length of the lift task is longer
than that of the push task. Since the score is the unnormalised
cumulative reward over all time steps, this means that the scores for
the lift task tend to be higher than those of the push tasks. This
results in the lift task getting a higher weight in the total score
computation, which is, however, intended as it is the more challenging
task.

\section{Data Collection} %
\label{sec:data_collection}

We collected datasets on 6 real TriFinger robots and in a simulated PyBullet environment \citep{trifinger-simulation} which was closely modeled after the real system. 
We used policies from \citet{guertler2023benchmarking} which were trained with Proximal Policy Optimization \citep{schulman2017proximal}. The training pipeline of \citet{guertler2023benchmarking} builds upon the work of \citet{Allshire21}, which uses a fast, GPU-based rigid body physics simulator \citep{makoviychuk2021isaac} to parallelize rollouts. In addition to the converged expert policies, we also consider an early training checkpoint with additive noise on the actions to which we refer as \emph{weak} policy.

We provided datasets with two different compositions for each of the two tasks: (i) The \emph{expert} dataset consists solely of trajectories collected with the converged policies and tests the ability to imitate a proficient behavior policy. (ii) In contrast to this, half of the \emph{mixed} dataset consists of trajectories obtained with the weak policy while the other half contains expert trajectories. Learning a good policy from this dataset requires a training algorithm that either performs credit assignment or distills high-quality trajectories to imitate. \tabref{tab:overview-datasets} summarizes the six datasets used in the competition.

\section{Methods}
\label{sec:methods}
We present the participants' solutions in the context of state-of-the-art offline RL methods to which we compare quantitatively in section \ref{sec:results}.

\subsection{State-of-the-Art Algorithms} %

In the following, we briefly summarize a selection of offline RL algorithms: Behavioral Cloning (BC)~\citep{BS95, Pomerleau91, RossGB11, TorabiWS18} is a purely supervised method in which the mean squared error between the actions of the behavioral policy $a \sim \pi_{\beta}(\cdot \mid s)$ and the learned policy $a \sim \pi(\cdot \mid s)$ is minimized. 
Critic Regularized Regression (CRR)~\citep{NZMSRSSGHF20} is a BC variant in which the actions are \textit{weighted} according to advantage-based estimates using Q-values. CRR optimizes the following objective:
\[
\argmax_{\pi}\mathbb{E}_{(s,a)\sim\mathcal{B}}\left[ f(Q_{\theta}, \pi, s, a) \log \pi(s, a)\right],
\]
where \( f \) is a non-negative, scalar function whose value is monotonically increasing in \( Q_{\theta} \). 
Advantage Weighted Actor Critic (AWAC)~\citep{NDGL20} is an actor-critic method in which the policy improvement step is formulated as a constrained optimization problem that forces the policy to stay close to the behavioral policy.
Conservative Q-Learning (CQL)~\citep{KumarZTL20} is an actor-critic method that combines the Bellman update in the critic loss with a conservative loss that aims to push down Q-values of out-of-distribution (OOD) actions. 
Implicit Q-Learning (IQL)~\citep{KNL21} is an offline RL algorithm that avoids out-of-distribution action queries during training. It mitigates overshooting of the value function by estimating a Q-function expectile, and then performs policy extraction with weighted Behavioral Cloning.
For more details on Offline RL, we refer the reader to the surveys~\cite{LKTF20,PMC22}.

\subsection{Team ``excludedrice'' (1st place)} %
\textbf{Qiang Wang and Robert McCarthy} -- University College Dublin
\vspace{0.2cm}

The training of the robot controller used by team \textit{excludedrice} is based on BC. Their full solution can be found at \citep{excludedrice-solution}. Simply put, in their work, they found that BC performed better when cloning expert demonstrations than when training with complex offline reinforcement learning applied to data containing demonstrations with mixed skill levels. Nevertheless, BC tends to perform poorly on mixed datasets that contain mixed skill levels, which can introduce ambiguity and make it more difficult for BC's supervised learning process to accurately perform the required regression. In general, BC is best suited to situations where the actions being modeled are conditioned on states from a unimodal distribution, or when the target action mode makes up the majority of the data in the dataset. After investigating the composition of the mixed quality datasets, they discovered that half of the data was collected by experts, and this subset of data was potentially adequate for training a good policy. Thus, their objective was to filter out this expert data for training a controller using BC. However, simple manual methods were unable to differentiate the expert data as the performance of both expert and non-expert data was similar. As a result, they proposed a novel semi-supervised learning data filtering approach in their strategy. Initially, they extracted a small portion of the data with the highest scores from the entire dataset that was presumed to be mostly collected by an expert agent. It is worth noting that the size of this initial extracted data subset was insufficient to train a well-performing policy model. They fed this portion of data into a neural network (NN) to learn patterns from expert data, and then used it as a binary classifier to separate out more expert data for training the next iteration of the NN. They repeated this semi-supervised learning process iteratively until the number of separated expert data no longer increased. They improved this algorithm after the competition, and their methodology can be found in \citet{wang2023:bd}.

Furthermore, they augmented the training data of the robotic arena using spatial rotation transformations, taking advantage of the rotational symmetry of the physical TriFinger robot. However, it is important to note that data augmented through mathematical theory may not be entirely accurate in real-world scenarios. For instance, factors like friction and the dimensions of different fingers of the robot may not be identical due to physical errors during manufacturing. To address this issue, they proposed a policy training paradigm to make the model trained on theoretical data to better fit the data distribution of the real robotic environment.

\subsection{Team ``decimalcurlew'' (2nd place)} %
\textbf{Hangyeol Kim and Jongchan Baek} -- Pohang University of Science and Technology (POSTECH)\\
\textbf{Wookyong Kwon} -- Electronics and Telecommunications Research Institute (ETRI)
\vspace{0.2cm}

Team \textit{decimalcurlew} used offline RL and a regularization technique to obtain robust policies that perform well in a real robot system, even with measurement noise. 
They trained a feed-forward neural network policy for each task with nonlinear rectified linear units and 400 and 300 hidden units.
The team preprocessed the dataset for training by performing state and action normalization, and scaling the actions to fit within the range of $[-1, 1]$.
They then trained the policy networks on the normalized dataset using the offline RL algorithm TD3+BC~\cite{FujimotoG21}.

Given the uncertainty of observation noise in a real-robot system, the team aimed to obtain policies that could work robustly against such noise. 
To accomplish this, they adapted the policy training objective of TD3+BC and added a regularization term that encourages the policy’s actions to be spatially smooth~\cite{MysoreM0S21}, leading to similar actions for comparable states in the robot system.
In the final stage of the challenge, their policies exhibited competitive performance across all tasks.

\subsection{Team ``superiordinosaur'' (shared 3rd place)} %
\textbf{Shanliang Qian} -- Independent
\vspace{0.2cm}

Team \textit{superiordinosaur} observed an important feature of the real dataset,
the vision tracking system is far from perfect and sometimes suffers from: i) high delay and ii) noisy cube pose estimation.
They proposed a simple approach that combines supervised learning, early stopping and the introduction of a validity check with a smoothing process that maintains a moving average of the cube pose.
In particular, for a new cube pose, their algorithm checks whether the delay is less than or the confidence is greater than certain thresholds, in order to decide whether to update the cube pose with a moving average or to keep the last cube pose instead.
The team also report unsatisfactory results with the TD3+BC algorithm and LSTM architectures.\\

\subsection{Team ``jealousjaguar'' (shared 3rd place)}
\textbf{Yasunori Toshimitsu, Mike Yan Michelis, Amirhossein Kazemipour,}\\
\textbf{Arman Raayatsanati, Hehui Zheng, and Barnabas Gavin Cangan} -- ETH Zurich
\vspace{0.2cm}

Team \textit{jealousjaguar} used the offline RL algorithm ``Implicit Q-Learning'' (IQL)~\cite{KNL21}, due to its ability to avoid out-of-distribution action queries during training and to mitigate a value function overshooting by estimating a Q-function expectile via an asymmetric $\ell_2$ loss.
They selected the IQL implementation provided by the open source library \textsc{d3rlpy}~\citep{SI21}.

The team has furthermore created an automation pipeline that allows submissions to be queued, and sent to the real robot cluster automatically (the RRC system only accepted submissions if there was no ongoing submission by the same team), which allowed the performance of the policy to be gauged periodically during training and uploaded to Weights \& Biases~\citep{wandb}.
This made the comparison of different algorithms and parameters easier, allowing the team to develop their own ideas and compare in real time with others.
They also introduced various methods to improve the performance and consistency of a policy, such as data augmentation, or the inclusion of previous observations and actions in the state.

\section{Results}
\label{sec:results}

We present the results of the real-robot stage in this section and compare them to what state-of-the-art offline RL algorithms can achieve on the challenge datasets. The results of the pre-stage are summarized in Appendix \ref{apd:additional-results}.

\subsection{Usage Statistics} %

\definecolor{superiordinosaur}{RGB}{142, 176,  49}
\definecolor{decimalcurlew}{RGB}{224, 155,  36}
\definecolor{jealousjaguar}{RGB}{235,  98,  53}
\definecolor{excludedrice}{RGB}{93, 129, 181}

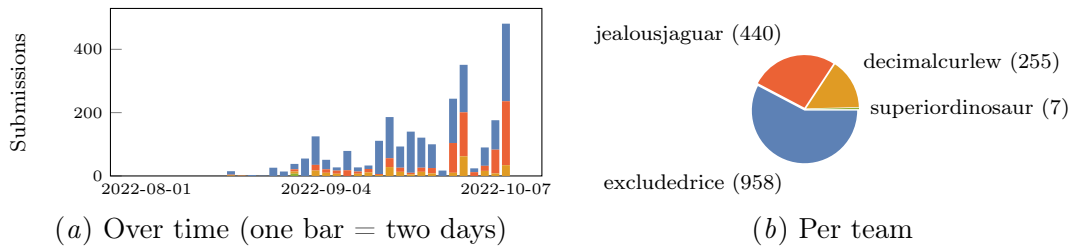
\begin{figure*}
    \centering
    \subfigure[Over time (one bar = two days)]{
    \centering
        \begin{tikzpicture}
          \scriptsize
          \begin{axis}[
              ybar stacked,
              height=1.5in,
              bar width=3pt,
              x=4pt,
              ymin=0,
              xtick=data,
              table/col sep=comma,
              legend style={cells={anchor=west}, legend pos=north west},
              reverse legend=true,
              xticklabels from table={rrc2022_hist_compressed.txt}{Date},
              xticklabel style={font=\tiny,}, %
              yticklabel style={font=\tiny},
              ylabel={Submissions},
              xtick pos=left,
              xtick style={draw=none},
              ytick pos=left,
            ]
            \addplot [draw=none, fill=superiordinosaur] table [y=superiordinosaur, meta=Date, x expr=\coordindex] {rrc2022_hist_compressed.txt};
            \addlegendentry{superiordinosaur}
            \addplot [draw=none, fill=decimalcurlew] table [y=decimalcurlew, meta=Date, x expr=\coordindex] {rrc2022_hist_compressed.txt};
            \addlegendentry{decimalcurlew}
            \addplot [draw=none, fill=jealousjaguar] table [y=jealousjaguar, meta=Date, x expr=\coordindex] {rrc2022_hist_compressed.txt};
            \addlegendentry{jealousjaguar}
            \addplot [draw=none, fill=excludedrice] table [y=excludedrice, meta=Date, x expr=\coordindex] {rrc2022_hist_compressed.txt};
            \addlegendentry{excludedrice}
            \legend{} %
          \end{axis}
        \end{tikzpicture} 
    }
    \subfigure[Per team]{
    \centering
        \begin{tikzpicture}
          \scriptsize
          \tikzset{
            lines/.style={draw=none},
          }
          \pie[hide number, sum=auto, radius=0.7, explode=0.02,
            color={superiordinosaur, decimalcurlew, jealousjaguar, excludedrice},
            style={lines},
          ]{7/superiordinosaur (7), 255/decimalcurlew (255), 440/jealousjaguar (440), 958/excludedrice (958)}
        \end{tikzpicture}
    }
    \caption{Number of job submissions to our robots over time and per team.}
    \label{fig:submission_statistics}
\end{figure*}

Throughout the real-robot stage, the participating teams submitted a total of 1660 jobs to the robots.  Moreover, the number of submitted jobs differed highly between teams.  Figure~\ref{fig:submission_statistics} shows the distribution of jobs over time and teams.

\subsection{Results} %

\tabref{tab:results} shows the average returns the teams achieved on all task/dataset combinations in the real-robot phase. The last column contains the overall score which is obtained by averaging these returns. We furthermore include the relevant benchmarking results from \citet{guertler2023benchmarking} as a point of reference\footnote{Note that the Mixed datasets correspond to the Weak\&Expert datasets in \citet{guertler2023benchmarking}.}.  The scores of the teams are compared to those of the benchmarked offline RL algorithms in \figref{fig:score-bars}. We additionally provide success rates in \tabref{tab:results:success_rates} in appendix \ref{apd:additional-results} .

Team excludedrice achieved the highest score by a significant margin by combining self-supervised dataset filtering with Behavioral Cloning. This approach even outperformed the behavior policy on the challenging mixed data from the Lift task, unlike all other competitors. As a result, excludedrice is the only team that exceeds the score of the behavior policies.

On the second rank, decimalcurlew also achieves good results on the Lift-Mixed dataset with a regularized version of TD3+BC. In contrast to this, the remaining teams fall behind on this decisive dataset. Surprisingly, the BC-based approach of team superiordinosaur slightly outperforms team jealousjaguar's solution built around the offline RL algorithm IQL. This may be a result of team superiordinosaur's effort to take the confidence of the tracking system into account when updating the pose estimate which seems to result in better performance on the Lift-Mixed dataset.

Two of the offline RL algorithms benchmarked in \citet{guertler2023benchmarking} achieved scores comparable to some of the top submissions. CRR reaches a score similar to that of decimalcurlew while IQL matches the score of superiordinosaur. Note, however, that the hyperparameters of the benchmarked algorithms used for the Lift task were optimized on the simulated version of Lift-Mixed. This requires a significant amount of computational resources but increases the scores on the real Lift-Mixed dataset.

\begin{table}
  \centering
  \caption{\textbf{Returns and overall score in the real-robot stage:} For each combination of task and behavior policy the mean return and the standard error of the mean return are given (failed runs correspond to a return of 0). The overall score is the return averaged over all tasks.}
  \label{tab:results}\vspace{0em}
  \begin{tabular}{@{}c@{ }r|ccccc@{}}
    \toprule 
     &  & Push/Expert & Push/Mixed & Lift/Expert & Lift/Mixed & Score \\
    \midrule
     & behavior policies &	\(660 \pm \ \, 2\) & \(429 \pm \ \,  4\) & \(\!\!\! 1064 \pm \ \, 7\) & \(851 \pm \ \, 8\) & \(751 \pm \ \, 3\) \\
    \midrule
    \multicolumn{2}{c}{\textbf{Teams}} \\
    \midrule
    1. & excludedrice &	\(624 \pm \ \, 6\) & \!\(\bf 635 \pm \ \,  5\) &	\!\(\bf 956 \pm 21\) & \!\(\bf 923 \pm 21\) &	\!\(\bf 784 \pm \ \,  8\) \\
	2. & decimalcurlew & \!\(\bf 639 \pm \ \, 4\) & \(613 \pm \ \,  5\) & \(841 \pm 20\) & \(717 \pm 18\) & \(703 \pm \ \,  7\) \\
    \multirow{2}{*}{3.} & superiordinosaur & \(618 \pm \ \, 6\) & \(575 \pm \ \,  8\) & \(856 \pm 22\) & \(571 \pm 17\) & \(655 \pm \ \,  7\) \\
     & jealousjaguar & \!\(\bf 639 \pm \ \, 5\) & \(561 \pm \ \,  7\) & \(855 \pm 19\) & \(506 \pm 17\) & \(640 \pm \ \,  7\) \\	 
    \midrule
    \multicolumn{2}{c}{\textbf{Algorithms}} &
    \multicolumn{5}{c}{(results from \citet{guertler2023benchmarking})} \\
    \midrule
    & BC & \(562 \pm 14\) & \(388 \pm 21\) & \(676 \pm 35\)  & \(437 \pm 26\) & \(516 \pm 13\) \\
    & CRR & \(638 \pm \ \, 8\) & \(621 \pm 11\) & \(890 \pm 34\)  & \(707 \pm 28\) & \(714 \pm 12\) \\
    & AWAC & \(623 \pm \ \, 9\) & \(567 \pm 14\) & \(747 \pm 36\)  & \(481 \pm 28\) & \(605 \pm 12\) \\
    & CQL & \(514 \pm 15\) & \(346 \pm 17\) & \(288 \pm 16\)  & \(269 \pm 14\) & \(354 \pm \ \, 8\) \\
    & IQL & \(592 \pm 10\) &  \(555 \pm 14\) & \(900 \pm 32\)  & \(574 \pm 31\) & \(655 \pm 12\) \\
    \bottomrule
  \end{tabular}
\end{table}

\begin{figure}
    \centering
    \includegraphics[width=0.85\textwidth]{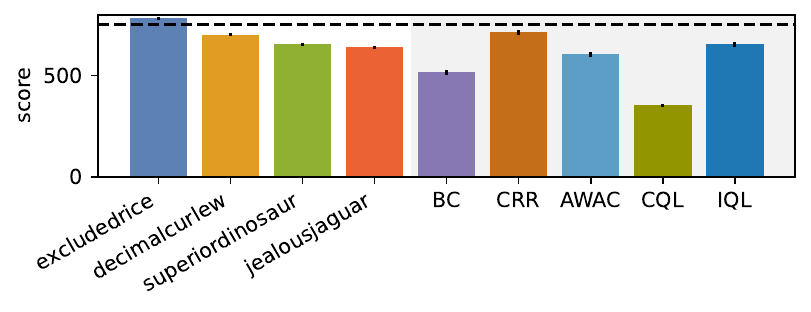}
    \caption{\textbf{Scores in the real-robot stage:} Overall scores of the winning teams and state-of-the-art offline RL algorithms for comparison. The averaged score of the behavior policies is indicated by a dashed line.}
    \label{fig:score-bars}
\end{figure}

\subsection{Challenge Takeaways} %
\label{sec:takeaways}

{\parindent0pt\paragraph{Imitation learning vs offline RL:}}In principle, offline RL should be able to outperform imitation learning on datasets containing suboptimal trajectories (which is the case for all challenge datasets), as it takes the reward signal into account. We were therefore surprised that two out of four top teams built their methods around Behavioral Cloning. We see several factors that could contribute to the popularity of BC: (i) Offline RL algorithms are sometimes difficult to implement and tune, unlike BC, (ii) the robotics community might not have fully adopted the offline RL paradigm yet, and (iii) in practice other parts of the method might have a bigger impact on performance.
\medskip
{\parindent0pt\paragraph{Algorithmic vs. problem-specific adaptations:}}Instead of optimizing the choice of learning algorithm or the algorithm itself, most teams concentrated on orthogonal contributions like filtering, data augmentation and regularization which were partly tailored to the robotics problem. They reported significant improvements in performance following this strategy.
\medskip
{\parindent0pt\paragraph{Simulation vs. real world:}} The gap between expert policy and learned policies was, on average, bigger on real-world data. This may be caused by more complex real-world dynamics as also discussed in \citet{guertler2023benchmarking}.

\section{Conclusion} %

In summary, the competition provided an opportunity to apply offline RL where it matters most: in the real world. 
The results show somewhat surprisingly that simpler methods, such as Behavioral Cloning combined with suitable filtering and data augmentation, can be more effective in real-world applications than more elaborate offline RL algorithms. This in turn means that more research on offline RL is required and that new algorithms should be also evaluated on real hardware \citep{guertler2023benchmarking}.   

\acks{
We thank
Thomas Steinbrenner for the maintenance of the robot platforms. We are furthermore grateful for the financial support of DeepMind that made the prize money possible.

GM and BS are members of the Machine Learning Cluster of Excellence, EXC number 2064/1 – Project number 390727645. This work was supported by the Volkswagen Stiftung (No 98 571). We acknowledge the support from the German Federal Ministry of Education and Research (BMBF) through the Tübingen AI Center (FKZ: 01IS18039B).
}

\bibliography{references.bib}

\appendix
\FloatBarrier
\section{Datasets}

We provide an overview of the datasets provided to the participants in table \ref{tab:overview-datasets}.

\begin{table}[h!]
\caption{Overview of the offline RL datasets provided to the participants.}
    \label{tab:overview-datasets}
    \begin{adjustbox}{max width=\textwidth}
    \begin{tabular}{@{}l|l|cccc@{}}
    \toprule
    task & dataset & overall duration [h] & \#episodes & \#transitions [$10^6$] & episode length [s]\\
    \midrule
    \multirow{3}{*}{Push-} 
    & \DPushSimExp & 16 & 3840 & 2.8 & 15  \\
    & \DPushRealExp  & 16 & 3840 & 2.8 & 15  \\
    & Real-Mixed & 16 & 3840 & 2.8 & 15  \\
    \midrule
    \multirow{3}{*}{Lift-} 
    & \DLiftSimExp & 20 & 2400 & 3.6 & 30  \\
    & \DLiftRealExp & 20 & 2400 & 3.6 & 30  \\
    & Real-Mixed & 20 & 2400 & 3.6 & 30  \\
    \bottomrule
    \end{tabular}
    \end{adjustbox}
    
\end{table}

\section{Complete set of rules}\label{apd:rules}

The complete set of rules of the Real Robot Challenge 2022 was as follows:

\begin{itemize}  
    \item Any algorithmic approach may be applied that learns the behavior only from the provided data and does not make use of any hard-coded/engineered behavior. As an example, two prominent algorithmic approaches meeting this criteria are: offline reinforcement learning and imitation learning.
    \item It is not permitted to use data collected during evaluation rollouts or obtained from other sources.
    \item It is not permitted to use data provided for one task to train a policy for an other task (e.g. use simulation data for the real robot or the "expert" dataset for the "mixed" task).
    \item It is not permitted to filter the datasets based on the position of a sample in the dataset. However, you may filter based on the properties of a transition or an episode if you want to.
    \item Participants may participate alone or in teams.
    \item Individuals are not allowed to participate in multiple teams.
    \item Each team needs to nominate a contact person and provide an email address through which they can be reached.
    \item Cash prizes will be paid out to an account specified by the contact person of each team. It is the responsibility of the team's contact person to distribute the prize money according to their team-internal agreements.
    \item To be eligible to win prizes, participants agree to release their code under an OSI-approved license and to publish a report describing their method in a publicly accessible way (e.g. on arXiv).
    \item Participants may not alter parameters of the simulation (e.g. the robot model) for the evaluation of the pre-stage.
    \item The organizers reserve the right to change the rules if doing so is absolutely necessary to resolve unforeseen problems. 
    \item  The organizers reserve the right to disqualify participants who are violating the rules or engage in scientific misconduct.
\end{itemize}

\section{Additional results}\label{apd:additional-results}

\begin{table}[ht]
  \centering
  \vspace{-1em}
  \caption{\textbf{Returns in the pre-stage} which were obtained by evaluating in the simulated environment after training on datasets recorded in simulation. The teams are listed as they ranked in the real-robot stage.}
  \label{tab:pre-stage-results}\vspace{.5em}
  \begin{tabular}{cr|cc}
    \toprule
     &  & Push/Expert &  Lift/Expert  \\
    \midrule
     & behavior policies &	674  & 1334 \\
    \midrule
    \multicolumn{2}{c}{\textbf{Teams}} \\
    \midrule
    & excludedrice & 676 & 1273 \\
	& decimalcurlew & 675 & 1132 \\
    & superiordinosaur & 653 & 1325 \\
	& jealousjaguar & 658 & 1137 \\	 
    \bottomrule
  \end{tabular}
\end{table}

\begin{table}[ht]
  \centering
  \vspace{-1em}
  \caption{\textbf{Success rates in the real-robot stage:} For each combination of task and behavior policy the mean success rate and the standard error of the mean success rate are given (failed runs correspond to a return of 0). The last column is the success rate averaged over all datasets.}
  \label{tab:results:success_rates}\vspace{.5em}
  \begin{tabular}{@{}c@{ }r|cccc@{}}
    \toprule
     &  & Push/Expert & Push/Mixed & Lift/Expert & Lift/Mixed \\
    \midrule
     & behavior policies &	\(0.92 \pm 0.01\) & \(0.51 \pm 0.01\) & \(0.66 \pm 0.01\) & \(0.40 \pm 0.01\)\\
    \midrule
    \multicolumn{2}{c}{\textbf{Teams}} \\
    \midrule
    1. & excludedrice & \( 0.82 \pm 0.02 \) & \( 0.83 \pm 0.01 \) & \( 0.48 \pm 0.02 \) & \(\bf  0.46 \pm 0.02 \)\\
    2. & decimalcurlew & \( 0.80 \pm 0.02 \) & \( 0.71 \pm 0.02 \) & \( 0.28 \pm 0.02 \) & \( 0.13 \pm 0.02 \)\\
    \multirow{2}{*}{3.} & superiordinosaur & \( 0.80 \pm 0.02 \) & \( 0.69 \pm 0.02 \) & \( 0.40 \pm 0.02 \) & \( 0.11 \pm 0.02 \)\\
    & jealousjaguar & \( 0.79 \pm 0.02 \) & \( 0.59 \pm 0.02 \) & \( 0.25 \pm 0.02 \) & \( 0.03 \pm 0.01 \)\\
    \midrule
    \multicolumn{2}{c}{\textbf{Algorithms}} &
    \multicolumn{4}{c}{(results by \citet{guertler2023benchmarking})} \\
    \midrule
    & BC & \(0.74\pm0.02\)  & \(0.48\pm0.03\) & \(0.28\pm0.02\) & \(0.09\pm0.02\) \\
    & CRR & \(\bf 0.87\pm0.03\) & \(\bf 0.84\pm0.04\) & \(\bf 0.54\pm0.04\) & \(0.29\pm0.04\) \\
    & AWAC & \(0.80\pm0.01\) & \(0.69\pm0.03\) & \(0.31\pm0.02\) & \(0.12\pm0.03\) \\
    & CQL & \(0.54\pm0.06\) & \(0.14\pm0.00\) & \(0.00\pm0.00\) & \(0.00\pm0.00\) \\
    & IQL & \(0.75\pm0.03\) & \(0.68\pm0.03\) & \(0.48\pm0.03\) & \(0.15\pm0.01\) \\
    \bottomrule
  \end{tabular}
\end{table}

\clearpage

\section{Reward function and success criterion}
\label{apd:reward-success}

The reward is obtained by applying the logistic kernel \begin{equation}
    k(x)=\left(b+2\right)\left(\exp(a\lVert x\rVert) + b + \exp(-a\lVert x\rVert)\right)^{-1}
\end{equation}
to the difference between desired and achieved position (for the Push task) or the desired and achieved corner points of the cube (for the Lift task). The parameters $a$ and $b$ control the length scale over which the reward decays and how sensitive it is for small distances $x$, respectively. 

We consider an episode successful if at its end the desired position is matched up to a tolerance of \SI{2}{cm} and the deviation from the desired orientation does not exceed \SI{22}{\deg}, similar to \citet{Allshire21} and \citet{guertler2023benchmarking}.

\section{Code repositories of the winning teams}

\begin{table}[h]
    \centering
    \begin{adjustbox}{max width=1.0\textwidth}
    \begin{tabular}{l|l}
    Teams & URL to repository \\
    \midrule
    excludedrice & \url{https://github.com/wq13552463699/Real-Robot-Challenge-2022.git} \\
	decimalcurlew & \url{https://github.com/paekgga/RRC2022Training
}\\
    superiordinosaur & \url{https://github.com/QianSL/rrc_solution} \\
	jealousjaguar & \url{https://github.com/QianSL/rrc_solution}\\	 
    \end{tabular}
    \end{adjustbox}
    \caption{URLs of the code repositories of the winning teams.}
    \label{tab:team-repositories}
\end{table}

\end{document}